\documentclass[conference]{IEEEtran}
\IEEEoverridecommandlockouts

\usepackage{cite}
\usepackage{amsmath,amssymb,amsfonts}
\usepackage{graphicx}
\usepackage{textcomp}
\usepackage{xcolor}
\usepackage{booktabs}
\usepackage{comment}
\usepackage[table]{xcolor}
\usepackage[hidelinks]{hyperref}
\graphicspath{{./}}

\title{PACT: A Pragmatic Prompting Framework for Expressive and Voice-Preserving Speech-to-Speech Translation}

\usepackage{pifont}

\newcommand{\pair}[2]{#1\,\ensuremath{\vert}\,#2}

\usepackage[table]{xcolor}
\usepackage{pifont}
\usepackage{makecell}
\usepackage{multirow}

\author{%
\IEEEauthorblockN{Yuchen Song\IEEEauthorrefmark{1}\IEEEauthorrefmark{2} \quad Satoshi Nakamura\IEEEauthorrefmark{1}}
\IEEEauthorblockA{\normalfont
\IEEEauthorrefmark{1}The Chinese University of Hong Kong, Shenzhen, China\\
\IEEEauthorrefmark{2}Shenzhen Loop Area Institute, China\\
\texttt{yuchensong1@link.cuhk.edu.cn, snakamura@cuhk.edu.cn}}
}

\begin{document}

\maketitle
\hypersetup{
  pdftitle={PACT: A Pragmatic Prompting Framework for Expressive and Voice-Preserving Speech-to-Speech Translation},
  pdfauthor={Yuchen Song and Satoshi Nakamura},
  pdfkeywords={speech-to-speech translation, expressive speech translation, pragmatic prompting, paralinguistic transfer, voice preservation, soft prompting}
}

\begin{abstract}
Speech-to-speech translation should preserve not only what is said, but also who says it and how it is said. Existing end-to-end S2ST, speech MLLM, and cascade systems can preserve content or timbre, but often struggle to realize emotion, tone, rhythm, pauses, and pragmatic attitude in the target language. We propose PACT, a target-space pragmatic prompting framework for expressive and voice-preserving S2ST. PACT extracts source pragmatic cues, projects them into the target-language expression space, calibrates the projected representation to a realizable target subspace, and converts it into continuous prompts for a frozen speech generator. On bidirectional Spanish--English TOPI evaluation, PACT achieves the best overall performance among evaluated systems across content fidelity, voice preservation, and expression preservation metrics.
\end{abstract}


\section{Introduction}

Speech-to-speech translation (S2ST) is a key technology for cross-lingual communication, real-time interpretation, and expressive media dubbing. For these applications, an S2ST system should preserve not only what is said, but also who says it and how it is said. Beyond semantic fidelity, recent work has increasingly studied expressive and voice-preserving S2ST, where the target speech should retain speaker identity and natural expressive delivery~\cite{jia2022translatotron,barrault2023seamless,huang2023holistic,gupta2025survey,uniss2025,unitbased2025}.

However, cross-lingual expression preservation is not simply a matter of preserving source-side acoustic or style cues. In this work, we use pragmatic cues to refer to paralinguistic cues in speech that help listeners perceive communicative intent, such as affective stance, emphasis, rhythm, pausing, and conversational attitude. The same intent may be realized differently across languages: urgency, sarcasm, or frustration in Spanish may require different pitch movement, pause placement, and emphasis patterns when naturally expressed in English~\cite{ward2024pragmatic,ward2022dral,suzuki2022representing}. This target-language realization gap has also been observed in expressive dubbing, cross-lingual prosody transfer, and phrase-level speech translation~\cite{crossvoice2024,dubs2st2025,phraselevel2023}.

Existing end-to-end S2ST, speech MLLM, and cascaded voice/style-cloning systems can preserve content, timbre, or naturalness, but they often lack an explicit mechanism for mapping how source expression should be realized in the target language. This motivates a target-language expression signal for expressive and voice-preserving S2ST.

We propose \textbf{PACT}, short for \emph{Pragmatic Alignment with Calibrated Target-space Prompting}, a framework for expressive and voice-preserving S2ST. PACT separates generation into three complementary channels: a semantic channel for target-language text, a speaker channel for voice identity, and a pragmatic prompting channel for utterance-level expression. The pragmatic channel maps source pragmatic cues into a target-language expression space, calibrates the projected representation to a realizable target subspace, and converts it into continuous prompts for a frozen speech generator.

We evaluate PACT on the bidirectional Spanish--English TOPI benchmark, which provides paired source speech and human target-language reenactments for measuring expressive S2ST. We compare PACT with end-to-end S2ST systems, speech MLLMs, and cascaded expressive TTS systems under a unified speech-output evaluation protocol.

The primary contributions of this paper are threefold:
\begin{itemize}
\item We formulate expressive and voice-preserving S2ST as target-language expression realization, and propose PACT, a target-space pragmatic prompting framework that separates semantic translation, speaker preservation, and utterance-level expression control.
\item We introduce realization-calibrated pragmatic prompting, which maps source pragmatic cues into a target-language expression space, constrains them to a realizable target subspace, and injects them as continuous prompts into a frozen speech generator.
\item We demonstrate state-of-the-art overall performance on the bidirectional Spanish--English TOPI benchmark, with consistent gains across content fidelity, voice preservation, and expression preservation metrics.
\end{itemize}

\begin{figure*}[t]
  \centering
  \includegraphics[width=\linewidth]{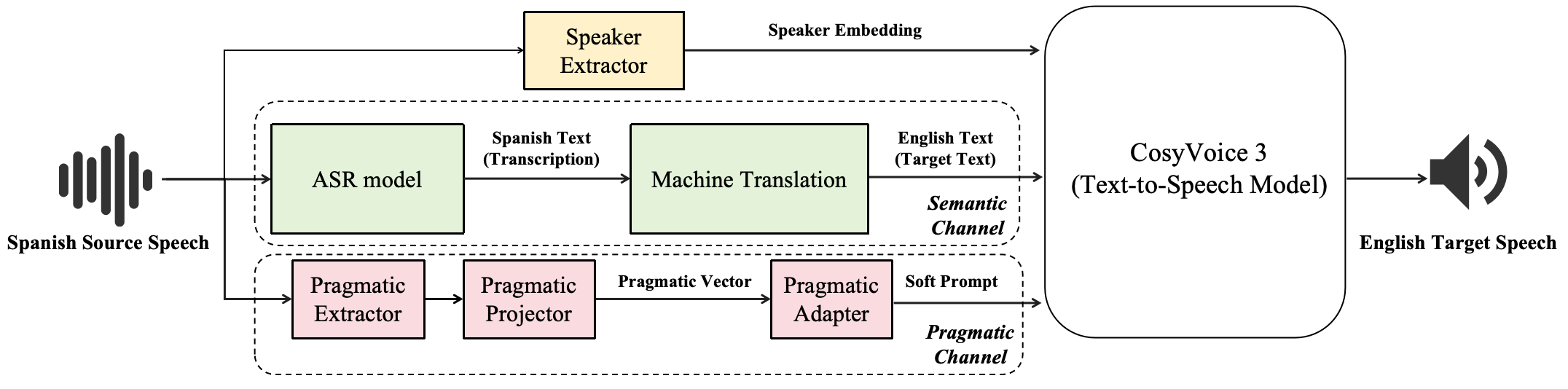}
  \caption{The overall architecture of PACT.}
  \label{fig:architecture}
\end{figure*}

\section{Related Work}
\label{sec:related_work}

\textbf{Expressive and voice-preserving S2ST.}
Recent S2ST systems increasingly aim to preserve not only linguistic content, but also speaker identity and expressive delivery. Translatotron-style and direct S2ST models study voice-preserving speech translation without relying entirely on text-only generation~\cite{jia2022translatotron,lee2022direct}. SeamlessExpressive extends multilingual S2ST toward vocal style and prosody preservation through prosody-aware unit generation and cross-lingual expressivity modeling~\cite{barrault2023seamless}. Recent systems such as UniSS further explore unified modeling for preserving timbre, emotion, and duration consistency~\cite{uniss2025}. Discrete-unit and style-transfer S2ST methods improve cross-lingual speech generation by modeling target speech units or additional acoustic/style representations~\cite{lee2022direct,popuri2022enhanced,wang2024speech}. These approaches demonstrate the importance of expressive S2ST, but expression transfer is typically handled implicitly through end-to-end generation, unit prediction, style embeddings, or source-speech prompting. In contrast, PACT explicitly formulates expression transfer as target-space pragmatic prompting. 

\textbf{Prompt-conditioned generation and speech MLLMs.}
Cascaded ASR/MT plus prompt-conditioned TTS systems provide a practical route to expressive S2ST by combining strong text translation with voice or style cloning. Fish-Speech uses reference-audio tokens for token-prompted in-context TTS~\cite{fishspeech2023}; OpenVoice supports tone-color cloning and flexible style control~\cite{qin2023openvoice}; and XTTS-v2 enables cross-lingual voice cloning from a short reference clip~\cite{casanova2024xtts}. Recent speech MLLM and live-translation systems, such as Qwen3.5-LiveTranslate, further expose speech-to-speech generation and voice-cloning capabilities~\cite{qwen2026livetranslate}. These systems provide strong generation backbones and voice/style prompts, but their public interfaces do not explicitly map source-language pragmatic cues into a target-language expression space. PACT fills this gap by keeping the speech generator frozen while projecting, calibrating, and prompting with target-space pragmatic representations.

\section{Proposed Method}
\label{sec:system}

We propose \textbf{PACT} (Pragmatic Alignment with Calibrated Target-space Prompting), a modular framework for expressive and voice-preserving speech-to-speech translation. Given a source utterance in language $\ell_s$ and a target language $\ell_t$, PACT separates generation into three complementary channels: a \textit{semantic channel} for translated linguistic content, a \textit{speaker channel} for voice identity preservation, and a \textit{pragmatic prompting channel} for utterance-level expression control. The three channels are fused only at synthesis time through a frozen speech generator, as illustrated in Figure~\ref{fig:architecture}.

Unlike approaches that rely primarily on source-side audio, speaker, or style conditioning, PACT maps source pragmatic cues into the target-language expression space before using them to guide speech generation. This design allows the generated target speech to preserve the source speaker's voice while faithfully realizing the source utterance's paralinguistic intent in the target language.

\subsection{Semantic Channel}
\label{subsec:semantic_channel}

The semantic channel produces the target-language text that conditions speech generation. We first transcribe the source-language speech using Whisper Large V3~\cite{radford2023robust}, and then translate the transcript into the target language using Qwen2.5-72B-Instruct~\cite{yang2024qwen}. This cascade is used solely to provide the linguistic content for synthesis; speaker identity and expressive control are handled by the speaker-conditioning and pragmatic prompting channels described below. For same-backbone ablations, we use the same cached translations across all variants, ensuring that differences in the generated speech can be attributed to the prompting and synthesis components rather than to changes in the translated text.

\subsection{Speaker Channel}
\label{subsec:speaker_channel}

To preserve the source speaker's voice identity, PACT uses the native source-audio speaker-conditioning interface of CosyVoice 3~\cite{du2024cosyvoice}. Given the source utterance $x_s$, the speaker channel extracts a speaker condition $v_s$, which is provided to the speech generator as a separate conditioning signal from the pragmatic prompt. This separation is important: the speaker channel captures relatively stable speaker identity and timbre, whereas the pragmatic prompting channel controls utterance-level paralinguistic realization in the target language.

\subsection{Pragmatic Prompting Channel}
\label{subsec:pragmatic_channel}

The pragmatic prompting channel is the core of PACT. It converts source-side pragmatic cues into a target-language prompt representation that can guide a frozen speech generator.

\paragraph{Language-specific pragmatic features}
We use a pre-trained HuBERT Large model~\cite{hsu2021hubert} as the acoustic representation backbone. Following Ward et al.~\cite{ward2024pragmatic}, we use language-specific selected HuBERT dimensions associated with pragmatic similarity judgments. The resulting vector should be understood as an acoustic-pragmatic representation: it captures speech cues that listeners use to judge pragmatic similarity, rather than a discrete symbolic label of the speaker's intent. Let
\begin{equation}
    p_s = \mathrm{Pool}\big(M_{\ell_s}(\mathrm{HuBERT}(x_s))\big) \in \mathbb{R}^{d_s}
\end{equation}
denote the source pragmatic vector, where $M_{\ell_s}$ selects the language-specific pragmatic dimensions and $\mathrm{Pool}(\cdot)$ denotes temporal average pooling. In our Spanish--English setting, $d_{\mathrm{es}}=101$ and $d_{\mathrm{en}}=103$. Thus, Spanish-to-English uses a $101 \rightarrow 103$ mapping, while English-to-Spanish uses a $103 \rightarrow 101$ mapping.

\paragraph{Target-space pragmatic projection}
Because the selected HuBERT dimensions are language-specific, directly injecting a source-language pragmatic vector into target-language synthesis is not appropriate. PACT first maps the source pragmatic vector into the target-language expression space:
\begin{equation}
    z_{\mathrm{base}} = f_{\ell_s \rightarrow \ell_t}(p_s) \in \mathbb{R}^{d_t},
    \label{eq:base_projection}
\end{equation}
where $f_{\ell_s \rightarrow \ell_t}$ is a cross-lingual pragmatic projector fitted on paired DRAL training utterances~\cite{ward2022dral}. In the final PACT system, we use a data-efficient partial least squares (PLS) projector, which is well suited to the low-resource paired setting and explicitly models source--target covariance.

\paragraph{Target-space realization calibration}
The projected vector $z_{\mathrm{base}}$ may still contain directions that are difficult for a frozen speech generator to realize reliably. To constrain the projected cue to the target-language expression manifold, PACT applies a conservative realization calibration:
\begin{equation}
    z_{\mathrm{cal}}
    =
    \mu_t
    +
    \alpha \,
    U_{t,k} U_{t,k}^{\top}
    (z_{\mathrm{base}} - \mu_t),
    \label{eq:calibration}
\end{equation}
where $\mu_t$ is the mean target-language pragmatic vector computed from the training split, $U_{t,k}$ is an orthonormalized target-side PLS subspace, and $\alpha$ controls the strength of the calibrated residual. The subspace dimension $k$ and shrinkage coefficient $\alpha$ are selected on the development set and fixed before evaluation. This calibration preserves source-conditioned pragmatic variation while suppressing out-of-distribution directions that the frozen generator may not faithfully realize.

\paragraph{Soft prompt adapter}
The calibrated target-language pragmatic vector is converted into continuous prompt tokens by a lightweight adapter:
\begin{equation}
P_{\mathrm{soft}}
=
A_{\phi,t}(z_{\mathrm{cal}})
\in \mathbb{R}^{m \times d_{\mathrm{model}}},
\label{eq:adapter}
\end{equation}
where $m$ is the number of soft prompt tokens and $d_{\mathrm{model}}$ is the hidden dimension of the CosyVoice 3 language model. In our final system, $A_{\phi,t}$ is implemented as a lightweight MLP that maps $z_{\mathrm{cal}}$ to an $m d_{\mathrm{model}}$-dimensional vector, which is then reshaped into $m$ continuous prompt embeddings. We use a target-language adapter so that the prompt is expressed in the conditioning space appropriate for the target-language speech generator. Compared with discrete style labels or text-based emotion tags, continuous prompts provide a fine-grained control interface for acoustic and pragmatic variation without modifying the frozen generator.

\begin{figure}[t]
  \centering
  \includegraphics[width=\linewidth]{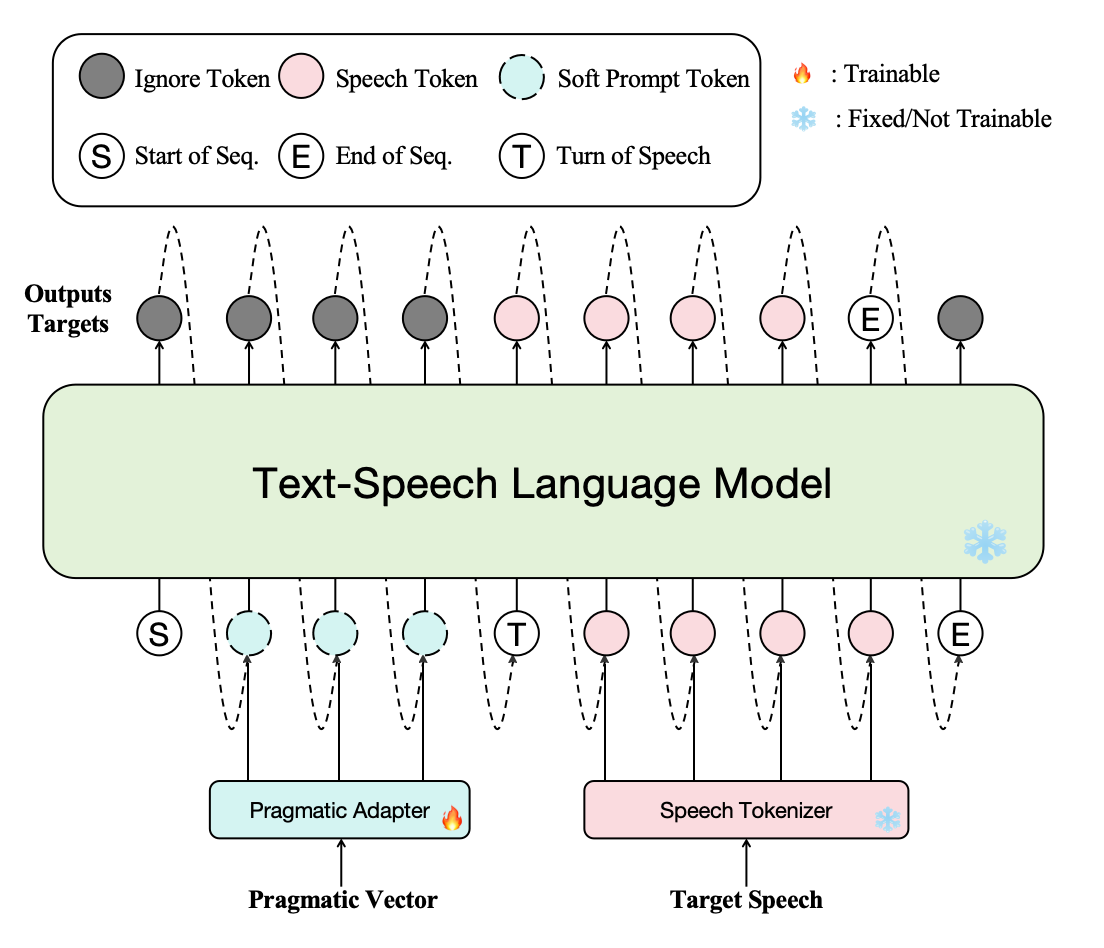}
  \caption{The training pipeline of the Pragmatic Adapter.}
  \label{fig:training_pipeline}
\end{figure}

\subsection{Training and Inference}
\label{subsec:training_inference}

PACT uses CosyVoice 3~\cite{du2024cosyvoice} as the target speech generator and keeps it \textbf{fully frozen} throughout training and inference. We do not fine-tune its language model, speech tokenizer, speaker-conditioning pathway, or waveform decoder. Instead, PACT estimates or trains only the components required to produce pragmatic soft prompts: the cross-lingual projector, the realization calibration statistics, and the pragmatic adapter. This makes the method parameter-efficient and preserves the generator's pre-trained speech synthesis capability, ensuring that performance gains come from target-space pragmatic prompting rather than from training a larger or domain-specific speech generator.

\paragraph{Projector and calibration fitting}
For each translation direction $\ell_s \rightarrow \ell_t$, we extract paired source- and target-language pragmatic vectors from the DRAL training split:
\begin{equation}
\left\{(p_s^{(i)}, p_t^{(i)})\right\}_{i=1}^{N}.
\end{equation}
The PLS projector $f_{\ell_s \rightarrow \ell_t}$ is fitted on these paired vectors. The target-language mean $\mu_t$ and the target-side realization subspace $U_{t,k}$ are also estimated only from the training split. No statistics from the evaluation split are used for projector fitting or calibration.

\paragraph{Adapter training}
The pragmatic adapter is trained to map target-language pragmatic vectors to soft prompt tokens for the frozen generator. Given a target-language training utterance with text $T_t$, target pragmatic vector $p_t$, and speech-token sequence $s_{1:N}$, the adapter produces
\begin{equation}
P_{\mathrm{soft}} = A_{\phi,t}(p_t).
\end{equation}
Only the adapter parameters $\phi$ are optimized, using the standard speech-token cross-entropy objective:
\begin{equation}
\mathcal{L}_{\mathrm{CE}}
=
-
\sum_{n=1}^{N}
\log
P_{\psi}
\left(
s_n
\mid
s_{<n},
T_t,
A_{\phi,t}(p_t)
\right).
\label{eq}
\end{equation}
where $\psi$ denotes the frozen parameters of the CosyVoice 3 language model. Figure~\ref{fig:training_pipeline} illustrates this adapter-training pipeline. During training, the adapter learns from target-language pragmatic vectors extracted from real target speech; at inference time, this vector is replaced by the calibrated prediction $z_{\mathrm{cal}}$ obtained from the source utterance. Thus, target reference speech is never used during inference.

\paragraph{Inference}
At inference time, PACT first computes the pragmatic soft prompt from the calibrated target-space vector:
\begin{equation}
P_{\mathrm{soft}} = A_{\phi,t}(z_{\mathrm{cal}}).
\end{equation}
The soft prompts are prepended to the target text-token embedding sequence as continuous conditioning prefixes:
\begin{equation}
\tilde{E}(T_t)
=
[P_{\mathrm{soft}}; E(T_t)],
\end{equation}
where $E(T_t)$ denotes the text-token embeddings. The CosyVoice 3 language model predicts speech tokens conditioned on the prompted text representation $\tilde{E}(T_t)$. The source speaker condition $v_s$ is then provided to the flow-matching decoder, which reconstructs the final waveform from the generated speech tokens while preserving the source speaker identity.

\section{Experiments}
\label{sec:experiments}

\subsection{Implementation and Training Details}
\label{subsec:implementation}

We split the DRAL training portion into 2,577 training pairs and 286 development pairs. Since the Emilia subset used here does not contain Spanish speech, the two directions use different adapter-training protocols. For Spanish-to-English, the English-target adapter is pre-trained on a 400-hour Emilia subset for 8 epochs and fine-tuned on DRAL for 30 epochs. For English-to-Spanish, the Spanish-target adapter is trained only on DRAL for 30 epochs. In both directions, only the pragmatic adapter is updated and the CosyVoice 3 generator remains frozen.

The pragmatic adapter is a lightweight MLP that maps a target-space pragmatic vector into seven 896-dimensional continuous prompt tokens. It uses a 512-dimensional hidden size with GELU activations and LayerNorm, and contains approximately 3.5M trainable parameters. For each direction, we fit a 64-component scaled PLS projector on paired pragmatic vectors from the DRAL training split, mapping 101$\rightarrow$103 dimensions for Spanish-to-English and 103$\rightarrow$101 for English-to-Spanish.

Target-space realization calibration is applied only in Spanish-to-English, where the English-target adapter benefits from Emilia pre-training. It uses the target-language training mean and the PLS target subspace, with $k=64$ and $\alpha=0.25$. For English-to-Spanish, we disable calibration based on development-set performance and use the PLS-projected Spanish vector directly, avoiding over-constraining the DRAL-only Spanish target-space prompt representation.

The adapter is optimized with Adam using a learning rate of $1\times10^{-5}$, 1,000 warm-up steps, gradient clipping of 5, and gradient accumulation of 2. The prompt length is selected from $m \in \{1,3,5,7,9,11,13,15\}$ on the development set, with $m=7$ used in the final system. All waveforms are generated at 24 kHz. Architecture choices, projector variants, calibration settings, and checkpoints are selected on the development set; the evaluation split is used only for final reporting. All training and inference are conducted on a single NVIDIA RTX 5090.

\providecommand{\pair}[2]{#1\,\ensuremath{\vert}\,#2}

\begin{table*}[t]
\centering
\caption{Bidirectional Spanish--English expressive S2ST evaluation on the TOPI benchmark. Each metric cell reports Es$\rightarrow$En $\vert$ En$\rightarrow$Es. All bounded metrics except ASR-BLEU and eG-Z are scaled by 100. The best and second-best results in each direction are in \textbf{bold} and \underline{underlined}, respectively. $\dagger$ denotes closed-source systems.}
\label{tab:main_results}
\small
\setlength{\tabcolsep}{2.5pt}
\resizebox{\textwidth}{!}{%
\begin{tabular}{cl|ccccccc}
\toprule
\textbf{Category}
& \textbf{System}
& \textbf{ASR-BLEU $\uparrow$}
& \textbf{Prag Sim. $\uparrow$}
& \textbf{Spk Sim. $\uparrow$}
& \textbf{SER-JSD $\downarrow$}
& \textbf{Emo Sim. $\uparrow$}
& \textbf{eG-Z $\downarrow$}
& \textbf{Dur. Ratio $\uparrow$} \\
\midrule
\multirow{4}{*}{\textit{S2ST}}
& SeamlessExpressive
& \pair{21.6}{23.9}
& \pair{72.0}{\underline{81.4}}
& \pair{\underline{66.3}}{\underline{70.6}}
& \pair{26.0}{13.9}
& \pair{72.6}{48.0}
& \pair{3.28}{2.81}
& \pair{69.3}{\underline{81.4}} \\

& SeamlessM4T-v2
& \pair{17.7}{10.7}
& \pair{58.5}{67.9}
& \pair{55.2}{61.6}
& \pair{44.8}{45.7}
& \pair{56.7}{24.9}
& \pair{3.24}{1.99}
& \pair{67.3}{73.7} \\

& Fairseq S2UT
& \pair{12.3}{1.7}
& \pair{50.7}{74.2}
& \pair{49.3}{55.2}
& \pair{56.9}{37.2}
& \pair{61.4}{45.6}
& \pair{3.25}{1.89}
& \pair{70.8}{74.9} \\

& Qwen3.5-LiveTranslate\textsuperscript{\textdagger}
& \pair{22.6}{21.1}
& \pair{67.7}{62.1}
& \pair{65.5}{57.0}
& \pair{21.5}{43.0}
& \pair{62.3}{48.1}
& \pair{2.83}{2.27}
& \pair{\underline{83.0}}{76.2} \\
\midrule

\multirow{2}{*}{\textit{MLLM}}
& Qwen3.5-Omni-Plus\textsuperscript{\textdagger}
& \pair{21.9}{\underline{27.3}}
& \pair{54.2}{64.3}
& \pair{59.5}{60.7}
& \pair{35.5}{44.9}
& \pair{33.2}{44.4}
& \pair{4.55}{2.37}
& \pair{74.8}{74.9} \\

& GPT-Audio-1.5\textsuperscript{\textdagger}
& \pair{21.0}{23.3}
& \pair{40.0}{55.7}
& \pair{48.5}{49.2}
& \pair{33.7}{24.7}
& \pair{63.2}{31.3}
& \pair{6.18}{7.00}
& \pair{66.8}{68.1} \\
\midrule

\multirow{3}{*}{\textit{Cascaded}}
& OpenVoice V2
& \pair{\underline{27.7}}{24.7}
& \pair{60.2}{53.5}
& \pair{63.1}{46.4}
& \pair{32.5}{41.4}
& \pair{\underline{73.4}}{40.5}
& \pair{\underline{2.37}}{2.65}
& \pair{77.5}{73.6} \\

& Fish-Speech
& \pair{26.6}{23.3}
& \pair{\underline{73.8}}{79.5}
& \pair{61.4}{69.6}
& \pair{\underline{17.7}}{\textbf{10.1}}
& \pair{60.5}{\underline{51.5}}
& \pair{2.46}{\underline{1.54}}
& \pair{67.6}{\textbf{82.5}} \\

& XTTS-v2
& \pair{26.4}{21.7}
& \pair{54.0}{61.5}
& \pair{58.3}{50.4}
& \pair{27.3}{36.6}
& \pair{70.2}{36.5}
& \pair{5.02}{2.79}
& \pair{74.4}{58.2} \\
\midrule

\rowcolor{gray!15}
\textit{Ours}
& \textbf{PACT}
& \pair{\textbf{30.7}}{\textbf{29.3}}
& \pair{\textbf{79.3}}{\textbf{82.9}}
& \pair{\textbf{72.9}}{\textbf{72.5}}
& \pair{\textbf{8.9}}{\underline{12.4}}
& \pair{\textbf{75.3}}{\textbf{54.2}}
& \pair{\textbf{1.20}}{\textbf{1.49}}
& \pair{\textbf{84.2}}{77.7} \\
\bottomrule
\end{tabular}%
}
\end{table*}

\subsection{Benchmark and Evaluation Protocol}
\label{subsec:setup}

We evaluate PACT on the released TOPI Spanish--English benchmark~\cite{topi2026challenge, ward2022dral}. The benchmark consists of paired Spanish--English utterances that are closely matched for communicative intent: bilingual speakers first participate in spontaneous conversations and then re-create selected utterances in the other language. Because spontaneous conversational speech naturally contains rich affective and pragmatic cues, this design makes the benchmark particularly suitable for expressive S2ST. The paired target-language reenactments provide human references for evaluating lexical content, speaker consistency, and target-language realization of paralinguistic and pragmatic intent. In our experiments, we use the 30 bilingual evaluation pairs for which target-language transcriptions are available, and evaluate both Spanish-to-English and English-to-Spanish directions.

\subsection{Compared Systems}
\label{subsec:baselines}

We compare PACT against three complementary baseline families. \textbf{End-to-end S2ST systems} include SeamlessExpressive and SeamlessM4T-v2~\cite{barrault2023seamless}, Fairseq S2UT~\cite{lee2022direct,popuri2022enhanced}, and Qwen3.5-LiveTranslate~\cite{qwen2026livetranslate}, covering expressive, unit-based, and voice-preserving S2ST paradigms. \textbf{Speech MLLM systems} include Qwen3.5-Omni-Plus~\cite{qwen2026omni} and GPT-Audio-1.5~\cite{openai2026audio}. \textbf{Cascaded systems} combine ASR/MT with OpenVoice V2~\cite{qin2023openvoice}, Fish-Speech / OpenAudio S1-mini~\cite{fishspeech2023}, or XTTS-v2~\cite{casanova2024xtts}. For PACT and all cascaded baselines, we use the same cached ASR/MT target text so that differences mainly reflect target-speech generation, speaker preservation, and expressive realization rather than upstream transcription or translation.

\subsection{Metrics}
\label{subsec:metrics}

We use a multi-dimensional target-referenced evaluation suite covering content fidelity, voice preservation, and expression preservation. Except for ASR-BLEU, all objective metrics are computed between the generated speech and the corresponding human target-language reference. For content fidelity, we transcribe generated speech with Whisper Large V3~\cite{radford2023robust} and compute BLEU~\cite{papineni2002bleu} against the human target-language reference transcription, denoted as ASR-BLEU. For voice preservation, we compute target-referenced speaker similarity (Spk Sim.) as the cosine similarity between WavLM-Large embeddings~\cite{chen2022wavlm} of generated and reference speech. This target-referenced protocol is appropriate for evaluation because the reference recordings are target-language reenactments of the same communicative intent, allowing us to evaluate target-language realization without relying on cross-lingual acoustic matching.

For expression preservation, we report the TOPI pragmatic similarity metric (Prag Sim.)~\cite{ward2024pragmatic,topi2026challenge}, computed between generated and reference speech. We also report complementary expressive-speech diagnostics: SER-JSD, the Jensen--Shannon divergence between SUPERB emotion-recognition probability distributions~\cite{yang2021superb}; Emo Sim., the cosine similarity between Emotion2Vec embeddings~\cite{ma2024emotion2vec}; eG-Z, the Z-normalized RMSE between eGeMAPS functionals~\cite{eyben2016gemaps} extracted with openSMILE~\cite{eyben2010opensmile}; and Dur. Ratio, defined as $\min(d_{\mathrm{gen}}/d_{\mathrm{ref}}, d_{\mathrm{ref}}/d_{\mathrm{gen}})$.

\subsection{Main Results}
\label{subsec:main_results}

Table~\ref{tab:main_results} presents the main bidirectional results on the TOPI Spanish--English benchmark. PACT establishes state-of-the-art performance among all evaluated systems, achieving the strongest overall performance across translation fidelity, target-language pragmatic realization, voice preservation, and expressive-speech diagnostics. In particular, PACT consistently leads on the three primary axes of expressive S2ST: ASR-BLEU, Pragmatic Similarity, and target-referenced Speaker Similarity in both translation directions.

\textbf{Balanced expressive S2ST performance.}
The main advantage of PACT is not limited to a single metric. Compared with all evaluated baselines, PACT substantially improves ASR-BLEU while also preserving stronger pragmatic and speaker similarity. This indicates that target-space pragmatic prompting does not trade off semantic fidelity for expressiveness; instead, it improves the final speech output while maintaining accurate target-language content.

\textbf{Target-language pragmatic realization.}
PACT's advantage is most evident when content fidelity, voice preservation, and expression preservation are considered jointly. Several baselines are competitive on individual expression-related metrics: SeamlessExpressive obtains a strong Pragmatic Similarity score in the English-to-Spanish direction, and Fish-Speech performs well on SER-JSD and duration matching. However, these systems do not achieve the same balance across ASR-BLEU, Pragmatic Similarity, Speaker Similarity, and complementary emotion/prosody diagnostics. This supports our central hypothesis that expressive S2ST benefits from an explicit target-language pragmatic control signal, rather than relying only on generic speech prompting or voice/style cloning.

\textbf{Comparison with cascaded and MLLM baselines.}
Cascaded systems such as Fish-Speech, OpenVoice V2, and XTTS-v2 benefit from strong ASR/MT and voice- or style-prompted synthesis, and they are competitive on individual metrics. Nevertheless, they remain less reliable in jointly preserving content, voice identity, and pragmatic intent. Speech MLLM systems produce fluent speech in many cases, but show weaker control over target-referenced speaker and expression similarity. These results demonstrate that PACT's explicit target-space pragmatic prompting provides a more effective control mechanism for expressive and voice-preserving S2ST.

\subsection{Subjective Evaluation}
\label{subsec:subjective}

We further conduct a subjective evaluation on the Spanish-to-English direction to complement the objective metrics. From the 30 evaluation pairs, we randomly select 10 utterances. For each utterance, the evaluation set contains the human English target reference recording and the English speech generated by five systems.

Five proficient English speakers rate each generated utterance on a 1-to-5 MOS scale along three dimensions: pragmatic similarity, speaker similarity, and naturalness. Pragmatic similarity measures whether the generated speech matches the reference in expression-related cues such as emotion, urgency, emphasis, rhythm, and conversational attitude. Speaker similarity measures whether the generated speech sounds like the same speaker as the English reference. Naturalness evaluates the generated speech itself, focusing on fluency, intelligibility, and the absence of audible artifacts.

\begin{table}[t]
\centering
\caption{Subjective MOS evaluation. Scores range from 1 to 5. The best and second-best scores are in \textbf{bold} and \underline{underlined}, respectively.}
\label{tab:subjective_results}
\small
\setlength{\tabcolsep}{4.5pt}
\resizebox{0.95\linewidth}{!}{%
\begin{tabular}{l|ccc}
\toprule
\textbf{System}
& \textbf{Prag Sim. $\uparrow$}
& \textbf{Spk Sim. $\uparrow$}
& \textbf{Naturalness $\uparrow$} \\
\midrule
Qwen3.5-LiveTranslate
& \underline{2.10}
& \underline{2.72}
& \underline{3.96} \\

Qwen3.5-Omni-Plus
& 1.64
& 1.80
& \textbf{4.78} \\

Fish-Speech
& 1.44
& 2.06
& 2.60 \\

SeamlessExpressive
& 1.58
& 2.14
& 1.80 \\
\midrule
\textbf{PACT}
& \textbf{3.42}
& \textbf{4.08}
& 3.64 \\
\bottomrule
\end{tabular}%
}
\end{table}

As shown in Table~\ref{tab:subjective_results}, PACT receives the highest subjective scores for both pragmatic similarity and speaker similarity. Across the subjective-evaluation samples, the objective Prag Sim. reported in Table~\ref{tab:main_results} correlates significantly with human-rated pragmatic similarity, with Pearson $r=0.413$ ($p=0.0029$) and Spearman $\rho=0.414$ ($p=0.0028$), supporting its use as the primary automatic pragmatic metric while motivating our complementary subjective and expressive-speech evaluations. Qwen3.5-Omni-Plus receives the highest naturalness score, consistent with its strong speech generation quality, but it is substantially weaker in target-referenced speaker similarity and pragmatic delivery. This result suggests that PACT's gains in objective pragmatic and speaker metrics correspond to perceptible improvements, while also showing that naturalness and expression preservation are not identical objectives.

\providecommand{\pair}[2]{#1,\ensuremath{\vert},#2}
\newcommand{\NA}{\makebox[1.8em][c]{--}}
\renewcommand{\pair}[2]{#1\,\ensuremath{\vert}\,#2}

\subsection{Ablation Study}
\label{subsec:ablation}

We conduct bidirectional same-backbone ablations to isolate the contribution of pragmatic prompting, target-space projection, realization calibration, and expressive pretraining. All variants use the same semantic channel, the same frozen CosyVoice 3 generator, the same speaker-conditioning interface, and the same decoding settings.

\textbf{Ablation configurations.}
\textit{Base} uses the default CosyVoice TTS generation prompt without any source-conditioned pragmatic prompt. \textit{Constant prompt} replaces PACT with a learned global soft prompt, independent of the input utterance; this tests whether improvements come merely from adding seven trainable prompt tokens. \textit{Source prompt} feeds the source-side pragmatic vector directly into the MLP adapter to generate soft prompts, without first projecting it into the target-language pragmatic space. \textit{Calibration variant} changes the calibration setting relative to the final PACT configuration: it removes realization calibration for Es$\rightarrow$En, where the final system uses calibration, and adds realization calibration for En$\rightarrow$Es, where the final system disables it based on development-set performance. \textit{w/o pretraining} removes Emilia pretraining for Es$\rightarrow$En; this ablation is not applicable to En$\rightarrow$Es because the Spanish-target adapter is already trained only on DRAL.

\textbf{Ablation results.}
Table~\ref{tab:ablation} shows that source-conditioned pragmatic prompting is essential in both directions. The Base and Constant prompt variants perform worse than PACT, indicating that neither the default TTS prompt nor a generic learned prompt provides sufficient utterance-specific pragmatic control. Source prompting improves over these controls, but remains below PACT in both ASR-BLEU and Pragmatic Similarity, supporting the need to map source pragmatic cues into the target-language expression space rather than injecting the source-side vector directly.

The calibration results further show that realization calibration should be applied direction-specifically. For Es$\rightarrow$En, removing calibration degrades ASR-BLEU, Pragmatic Similarity, and Speaker Similarity, suggesting that the projected English pragmatic vector benefits from being constrained to a realizable target subspace. In contrast, adding the same calibration to En$\rightarrow$Es reduces performance relative to the final ReversePLS configuration. This is likely because the Spanish-target adapter is trained only on DRAL without large-scale Spanish expressive pretraining; in this low-resource setting, calibration may over-constrain the target-space prompt representation. Finally, removing Emilia pretraining in Es$\rightarrow$En reduces all three metrics, confirming that large-scale expressive speech exposure helps the adapter learn a more stable interface to the frozen generator.

\begin{table}[t]
\centering
\caption{Bidirectional same-backbone ablation. Each metric cell reports Es$\rightarrow$En $\vert$ En$\rightarrow$Es. Prag Sim. and Spk Sim. are scaled by 100. The calibration variant reports w/o calibration for Es$\rightarrow$En and w/ calibration for En$\rightarrow$Es; \textemdash{} denotes not applicable. Best scores in each direction are in \textbf{bold}.}
\label{tab:ablation}
\small
\setlength{\tabcolsep}{4.2pt}
\resizebox{\linewidth}{!}{%
\begin{tabular}{lccc}
\toprule
\textbf{Variant}
& \textbf{ASR-BLEU $\uparrow$}
& \textbf{Prag Sim. $\uparrow$}
& \textbf{Spk Sim. $\uparrow$} \\
\midrule
Base
& \pair{27.0}{22.6}
& \pair{72.1}{80.3}
& \pair{69.5}{69.4} \\

Constant prompt
& \pair{27.6}{26.1}
& \pair{71.8}{80.8}
& \pair{64.0}{69.1} \\

Source prompt
& \pair{29.6}{27.4}
& \pair{77.4}{81.7}
& \pair{70.0}{71.1} \\

Calibration variant
& \pair{29.8}{25.3}
& \pair{78.9}{81.2}
& \pair{71.7}{70.9} \\

w/o pretraining
& \pair{29.6}{\NA}
& \pair{77.8}{\NA}
& \pair{70.9}{\NA} \\
\midrule
\rowcolor{gray!15}
\textbf{PACT}
& \pair{\textbf{30.7}}{\textbf{29.3}}
& \pair{\textbf{79.3}}{\textbf{82.9}}
& \pair{\textbf{72.9}}{\textbf{72.5}} \\
\bottomrule
\end{tabular}%
}
\end{table}

\section{Conclusion}
\label{sec:conclusion}

We presented PACT, a target-space pragmatic prompting framework for expressive and voice-preserving S2ST. PACT maps source pragmatic cues into a calibrated target-language expression space and injects them as continuous prompts into a frozen speech generator. Bidirectional TOPI results and ablations validate target-space projection, direction-specific calibration, and expressive adapter pretraining, with consistent gains in content, speaker, and pragmatic preservation.

\section*{Acknowledgment}
This paper is supported by Project W2531054 of the National Natural Science Foundation of China, and the Program for Guangdong Introducing Innovative and Entrepreneurial Teams. \textbf{AI-assisted writing disclosure:} OpenAI ChatGPT was used for language polishing and formatting assistance in the manuscript.

\clearpage
\bibliographystyle{IEEEtran}
\bibliography{mybib}

\end{document}